\documentclass[journal]{IEEEtran}

\usepackage{cite}
\usepackage{amsmath,amssymb}
\usepackage{graphicx}
\usepackage{booktabs}
\usepackage{tabularx}
\usepackage{array}
\usepackage{multirow}
\usepackage{makecell}
\usepackage{url}
\usepackage{xcolor}
\usepackage{placeins}
\usepackage{dblfloatfix}
\usepackage{microtype}
\usepackage{balance}

\newcolumntype{Y}{>{\raggedright\arraybackslash}X}
\newcommand{\Smax}{S_{\max}}
\newcommand{\Ssum}{S_{\Sigma}}
\newcommand{\Msum}{M_{\Sigma}}
\newcommand{\Mstep}{M_{\mathrm{step}}}
\newcommand{\Cterm}{C_{\mathrm{term}}}

\title{Decentralized Multi-Robot Task Allocation Under Degraded Communication: A Benchmark of Performance, Reliability, and Computation}

\author{James~Lott,~\IEEEmembership{Student Member,~IEEE,}
and~Vahraz~Honary,~\IEEEmembership{Senior Member,~IEEE}%
\thanks{J. Lott and V. Honary are with the Department of Electrical Engineering, University of San Diego, San Diego, CA 92110 USA (e-mail: jlott@sandiego.edu; vhonary@sandiego.edu).}}

\begin{document}
\maketitle

\begin{abstract}
Selecting a decentralized Multi-Robot Task Allocation (MRTA) method for embedded deployment on autonomous platforms requires considering more than route performance alone. We benchmark six decentralized MRTA allocators (CBAA, ACBBA, PI, HIPC, DMCHBA, and DGA) in the Collaborative Visit (CV) scenario to characterize tradeoffs among MinMax and MinSum travel, communication robustness and demand, allocation reliability, computational burden, and scale sensitivity. The core study uses 500 paired ten-target instances across 25 ideal and degraded communication conditions spanning Bernoulli loss, Gilbert--Elliott loss, and Rayleigh fading, with additional campaigns examining pre-allocation, execution-integrated computation, and sensitivity to grid size, robot density, and target load. Across the 24 impaired core conditions, DGA and DMCHBA achieved the lowest mean MinMax travel at 24.49 and 24.78 steps, respectively. HIPC narrowly led mean MinSum travel at 66.95 steps, followed by DGA at 67.22, with both methods occupying the top two in every impaired condition. DMCHBA had the lowest publication intensity at 2.08 publications per team step. In ten-target pre-allocation, HIPC and DMCHBA remained viable and stable in every tested condition, while ACBBA, PI, and DGA lost stability or viability as communication degraded. Under ideal delivery, median full-protocol computation $\Cterm$ in the primary ten-target comparison ranged from 4.88 ms for DMCHBA to 1.346 s for DGA. Static route quality preserved DGA and DMCHBA as the leading MinMax methods, while DGA led MinSum at three of four target loads and HIPC led at 50 targets. Static and execution-integrated computation rankings diverged as task load increased. The results identify distinct allocator operating regions across route objective, communication behavior, reliability, and computational constraints.
\end{abstract}


\section{Introduction}

Multi-robot systems support surveillance, exploration, and pick-and-delivery missions~\cite{2023quinton}, as well as environmental monitoring and search-and-rescue operations~\cite{2026abideen}. Coordinating these systems requires Multi-Robot Task Allocation (MRTA), which assigns mission tasks to members of a robot team~\cite{2004gerkey,2023quinton}. Recent surveys identify scalability, dynamic conditions, and uncertainty as persistent MRTA challenges~\cite{2026abideen}, while computational demand and the need to consider multiple performance objectives remain recurring concerns~\cite{2024athira}. Selecting a decentralized allocator is therefore not a one-dimensional route-cost decision because its suitability also depends on communication demand, allocation reliability, and the computation required before and during execution.

Collaborative Visit (CV) provides a controlled known-target setting for studying those tradeoffs~\cite{2020nayak,2025cao}. A team must visit a fixed set of target locations, and each robot may maintain an ordered route over several targets. MinMax minimizes the longest route assigned to any robot and therefore serves as a discrete proxy for mission completion time when robots share the same motion timing. In contrast, MinSum minimizes the total travel accumulated across the robot team~\cite{2025cao}. The objectives can favor different task distributions. MinMax rewards parallel completion, while MinSum can benefit from concentrating spatially compatible targets on fewer robots~\cite{lott2026}. Communication is integral to decentralized multi-robot coordination~\cite{2022gielis}, and lossy communication can degrade allocation quality~\cite{2020otte}. Evaluations of ACBBA under realistic lossy networks have further reported duplicate assignments and unassigned tasks~\cite{2018rantanen,2019rantanen}.

The closest cross-family benchmark by Nayak et al.\ compared CBAA, ACBBA, DHBA, HIPC, and PI under Bernoulli, Gilbert--Elliott, and Rayleigh fading~\cite{2020nayak}. It evaluated maximum-agent travel and maximum-agent message count over 50 randomized instances per condition, establishing that route and communication performance need not favor the same allocator. Cao et al.\ later compared CBAA, CBBA, HIPC, PI, DHBA, and DGA in pre-allocation and dynamic CV under packet loss, bit errors, and delay~\cite{2025cao}. Their study added allocation reliability, payload-aware communication overhead, and runtime efficiency to MinMax path quality using 100 Monte Carlo simulations per case at two agent--target scales.

Together, these studies established the importance of communication-aware comparison, but remained narrower in scope than the characterization pursued here. They did not provide a unified algorithm characterization spanning route objective, redundant work, workload concentration, allocation viability and stability, and computation accumulated before and during execution. They also did not include the newer multi-task DMCHBA allocator evaluated in this study or systematically isolate ranking continuity across independently varied workspace geometry, team size, and task load. The present work addresses these gaps through a paired CV framework with matched scenarios and statistical comparisons, in which six established non-learning allocators share the same robot shell, movement model, planner, safety layer, communication bus, and metric logger. Table~\ref{tab:prior} summarizes the relationship to the two closest predecessors.

\begin{table*}[!t]
\centering
\caption{Comparison with prior Collaborative Visit benchmarks.}
\label{tab:prior}
\scriptsize
\setlength{\tabcolsep}{3.5pt}
\renewcommand{\arraystretch}{1.10}
\begin{tabularx}{\textwidth}{@{}p{0.125\textwidth}YYY@{}}
\toprule
\textbf{Dimension} & \textbf{Nayak et al.~\cite{2020nayak}} & \textbf{Cao et al.~\cite{2025cao}} & \textbf{This work} \\
\midrule
Algorithms &
CBAA, ACBBA, DHBA, HIPC, PI &
CBAA, CBBA, DHBA, HIPC, PI, DGA &
CBAA, ACBBA, PI, HIPC, DMCHBA, DGA \\

Execution &
Repeated allocation and communication at a common 0.1-s cadence &
Synchronous pre-allocation and periodic asynchronous dynamic allocation &
Event-driven mission execution with separate strict-barrier pre-allocation and execution-timing studies \\

Communication &
Bernoulli, Gilbert--Elliott, and Rayleigh fading &
Packet loss, bit errors, and one-hop delay &
Bernoulli, Gilbert--Elliott, and Rayleigh fading \\

Design &
50 randomized trials per principal condition &
100 Monte Carlo trials at two agent--target scales &
500 paired core instances per condition with separate paired pre-allocation, execution-timing, and scale-sensitivity campaigns \\

Outcomes &
MinMax travel and maximum-agent messages &
MinMax path quality, allocation reliability, payload-aware overhead, and runtime &
MinMax and MinSum travel, duplicate work, workload concentration, communication demand, allocation reliability, static and execution-integrated computation, and scale continuity \\
\bottomrule
\end{tabularx}
\end{table*}

This work makes the following contributions:
\begin{enumerate}
\item A paired six-algorithm benchmark under ideal delivery and three degraded-communication families, using 112,500 unique core simulations across 500 matched instances and 25 communication conditions.
\item A joint characterization of realized MinMax and MinSum travel, publication demand, duplicate work, workload concentration, and degradation response.
\item A standardized pre-allocation protocol that separates viable allocation, stable allocation, full-protocol computation $\Cterm$, and achieved route quality over 4,500 trials.
\item A comparison of static protocol computation with computation and invocation frequency accumulated during asynchronous execution.
\item Complementary scale-sensitivity studies varying workspace geometry, robot density, and target load to test MinMax and MinSum ranking continuity and characterize diminishing returns from additional robots.
\end{enumerate}

The remainder of the paper is organized as follows: Section II reviews the relevant MRTA background and allocator families. Section III defines the Collaborative Visit problem and common execution and communication architecture. Section IV describes the six benchmarked allocation methods and planning-depth selection. Section V presents the experimental campaigns, evaluation metrics, and statistical analysis. Section VI reports and discusses the benchmark results, including communication and coordination behavior, pre-allocation reliability and computation, execution-integrated computation, scale sensitivity, and allocator-selection implications. Section VII concludes the paper.

\section{Background and Allocator Families}

\subsection{Multi-Robot Task Allocation}

MRTA assigns mission tasks to a robot team subject to robot, task, and environmental constraints. In this benchmark, each task is a known target location. A robot executes one target at a time, each target requires one robot, and the multi-task methods may maintain ordered sequences of future visits. Under established taxonomies, the problem is operationally ST--SR--TA with in-schedule dependencies because the cost of reaching a target can depend on its position within a route~\cite{2004gerkey,2013korsah}.

Centralized MRTA uses a common planner with team-wide information to compute assignments, which supports tightly coupled optimization but concentrates communication, computation, and decision authority at one node~\cite{2023quinton,2026abideen}. Decentralized MRTA distributes those decisions across the robots using local state and received peer information~\cite{2024athira,2022gielis}. This can improve resilience and responsiveness, although inconsistent local information may produce conflicting ownership, duplicate work, or suboptimal routes~\cite{2019rantanen}. The present study focuses on decentralized methods that continue allocating onboard without guaranteed agreement on the global mission state.

The evaluated suite spans single-task bid consensus, asynchronous bundle consensus, route-significance consensus, implicit local-team planning, deterministic matching, and evolutionary team-plan search. Several methods reason over multiple future assignments even though each robot physically executes only one target at a time.

\subsection{Consensus-Based Market Allocation}

Market-based methods assign tasks through locally computed utilities or costs and a winner-determination process~\cite{2023quinton}. CBAA combines single-task bidding with consensus on the winning bid for each task~\cite{2009choi}. CBBA extends this structure by greedily constructing an ordered task bundle whose values depend on the current path, while ACBBA adapts bundle consensus to asynchronous communication through timestamped deconfliction and selective rebroadcast rules~\cite{2010johnson}. Within the evaluated suite, CBAA represents single-task consensus and ACBBA represents asynchronous multi-task consensus.

Communication-constrained MRTA has been approached from several directions. Otte et al.\ examined auction-based task allocation under limited communication~\cite{2020otte}, while Bapat et al.\ developed distributed allocation methods intended to operate with very low communication demand~\cite{2022bapat}. Other work has treated communication as part of the allocation decision itself, including communication-aware metareasoning for selecting among allocation methods~\cite{2021carrillo}, consensus-based allocation adapted to constrained communication~\cite{2022raja}, and empirical evaluation under finite communication range~\cite{2025verma}. ACBBA-specific studies further showed that packet loss and increasing agent--task scale can produce duplicate or unassigned tasks when asynchronous deconfliction becomes unreliable~\cite{2018rantanen,2019rantanen}. Together, these studies motivate treating communication behavior as part of allocator performance rather than as an external implementation detail.

\subsection{Beyond Traditional Market Strategies}

PI and HIPC retain communication-mediated conflict resolution but depart from conventional bid competition. PI defines a task's significance from its contribution to route cost and alternates task inclusion with consensus and removal to improve the system objective~\cite{2016zhao}. HIPC combines implicit coordination, in which a robot plans for a locally modeled subset of the team, with plan consensus that resolves conflicting assignments. Its imperfect-situational-awareness extension detects inaccurate peer predictions and removes them when they impede convergence~\cite{2016johnson}.

PI has been extended for time-critical scheduling, distributed rescheduling, and deadline-constrained task assignment~\cite{2015whitbrook,2018turner,2024bai}. HIPC's prediction principle has also been used to suppress communication when neighboring robots can infer equivalent allocation information locally~\cite{2019kim}. These extensions emphasize specialized timing, heterogeneity, or communication objectives. The present benchmark instead places the established allocation mechanisms under matched CV execution and communication conditions.

\subsection{Deterministic and Evolutionary Optimization}

Optimization-based MRTA includes deterministic assignment solvers and population-based metaheuristics~\cite{2024athira,2026abideen}. DHBA distributes a one-to-one Hungarian assignment by allowing agents to reconstruct compatible cost information and independently solve the same assignment problem~\cite{2017ismail}. DMCHBA extends this lineage to multi-task assignment by cloning agents, adding pseudotasks when necessary, applying the Hungarian method once to the resulting square matrix, and ordering each robot's assigned targets through a local planner~\cite{2024samiei}. A recent extension incorporates explicit workload balancing into distributed Hungarian allocation~\cite{2024cao}.

DGA represents the evolutionary branch. Each robot maintains a population of complete team plans, continues improving that population during execution, and exchanges solution information that can be incorporated by peers~\cite{2020patel}. Related work has developed asynchronous collaborative genetic allocation for heterogeneous search-and-rescue teams~\cite{2021pallin}. DMCHBA and DGA therefore provide non-auction alternatives with different resource profiles. DMCHBA computes deterministic matching from a local cost matrix, whereas DGA repeatedly searches over complete team routes. Their different update structures motivate the static and execution-integrated computation analyses in this benchmark.

\section{System Design}

\subsection{Collaborative Visit Formulation}

Let $G\subset\mathbb{Z}^{2}$ denote the valid cells of a bounded discrete workspace, $R=\{r_1,\ldots,r_N\}$ the robot team, and $T=\{t_1,\ldots,t_M\}\subset G$ the known target cells. Robot $r_i$ starts at $p_i^0\in G$ and maintains an ordered local visit sequence
\begin{equation}
\pi_i=(t_{i,1},\ldots,t_{i,\ell_i}), \qquad t_{i,k}\in T.
\label{eq:route}
\end{equation}
A target is completed when any robot reaches its cell, and the mission ends when all targets have been visited. Robots do not return to their starting cells after all targets have been visited.

For four-connected motion, let $d_1(a,b)=\lVert a-b\rVert_1$ denote Manhattan distance. The nominal open-route length induced by $\pi_i$ is
\begin{equation}
L_i(\pi_i)=
\begin{cases}
0, & \ell_i=0,\\[1mm]
d_1(p_i^0,t_{i,1})+\displaystyle\sum_{k=1}^{\ell_i-1}d_1(t_{i,k},t_{i,k+1}), & \ell_i>0.
\end{cases}
\label{eq:routecost}
\end{equation}
A complete, conflict-free allocation $X=\{\pi_1,\ldots,\pi_N\}$ assigns every target exactly once. Its MinMax and MinSum objectives are
\begin{align}
J_{\max}(X)&=\max_{r_i\in R}L_i(\pi_i),\label{eq:minmax}\\
J_{\Sigma}(X)&=\sum_{r_i\in R}L_i(\pi_i).\label{eq:minsum}
\end{align}
Because all robots use the same motion-timing model, $J_{\max}$ is the nominal route-based makespan proxy and $J_{\Sigma}$ measures nominal aggregate travel. The complete allocation in (\ref{eq:minmax})--(\ref{eq:minsum}) is the route-quality object used in the pre-allocation study.

During asynchronous execution, each robot maintains a rolling local sequence over the currently active targets, and the union of those local sequences need not cover every active target at every event. The mission is completed through repeated allocation, movement, target completion, and route repair. The online benchmark reports realized movement, so its outcomes include travel introduced by online reallocation, duplicate target visits, and collision-safe detours.

Each robot removes a target after completing it locally or after receiving a state report that places a peer at that target. When completion information is lost, multiple robots may retain and visit the same target. Conflict-free local ownership and eventual target completion are therefore the desired online conditions, while duplicate visits characterize redundant work arising from inconsistent completion information.

\subsection{Asynchronous Robot Architecture}

At the execution level, the benchmark uses an event-driven discrete-grid simulator. Every trial contains a ground-truth world model, a shared robot-execution shell, and an algorithm-specific allocator for each robot. Although the world records target locations, robot visits, first completions, duplicate visits, and mission termination, allocators operate only from local mission state and delivered peer messages and cannot query another robot's internal state or the global completion record.

Path planning, motion, target-state updates, communication, collision safety, and metric logging are handled by the common robot shell. Trials advance through a priority queue of wake events in which due messages are delivered, one robot completes a shell-level action, and its next wake is scheduled. The allocator is called when the robot lacks a valid target or when an algorithm or recovery event clears the current assignment. Pending turns and movement actions continue without another call, allowing invocation frequency to emerge from each method's execution behavior.

The nominal condition places four robots on a four-connected, obstacle-free $19\times19$ grid. They begin at $(0,0)$, $(0,6)$, $(0,12)$, and $(0,18)$, facing east. Cardinal movement lasts $1.60$ s on average with $0.10$ s uniform jitter, clamped to $1.50$--$1.70$ s, and quarter turns require $0.30$ s. These values are modeled on the real-world testbed values as described in~\cite{mypaper1}. All methods use the same A* planner with a Manhattan heuristic, unit movement cost, a heading-change cost, and a penalty for re-entering locally visited cells.

Collision safety is enforced through a common intent layer. Before moving, a robot publishes its intended next cell and waits for the intent-settling interval. A move is blocked when another robot occupies or claims the conflicting cell, after which the shell replans, waits, or temporarily quarantines a repeatedly blocked target. Intent messages retain communication delay but bypass loss. They support physical safety and do not serve as allocator coordination messages.

\subsection{Communication Models}

Each publication is evaluated independently for every directed receiver $j\neq i$. A delivered copy incurs an end-to-end communication delay $\tau_c=\tau_0+\epsilon_c$, where $\tau_0=0.04$ s and $\epsilon_c\sim U[-0.01,0.01]$ s. One sender publication is counted once regardless of the number of receivers, while delivered and dropped directed copies are logged separately.

Independent loss is modeled by dropping each directed copy with configured probability $p_{\mathrm{drop}}$. For burst-correlated loss, every directed link maintains a Gilbert--Elliott process~\cite{1960gilbert,1963elliott}. With stationary drop fraction $p_{\mathrm{drop}}$ and fixed lag-one state correlation $\rho=0.8$, the state persistence probabilities are
\begin{equation}
p_{GG}=1-(1-\rho)p_{\mathrm{drop}}, \qquad p_{BB}=\rho+(1-\rho)p_{\mathrm{drop}}.
\label{eq:ge}
\end{equation}
Each link is initialized from its stationary distribution, delivery is evaluated from the current state, and the state then transitions.

Distance-dependent delivery is represented by a Rayleigh fading model that combines log-distance path loss with an exponential power gain~\cite{2003zheng}. For robot positions $p_i$ and $p_j$,
\begin{align}
d_m&=\max\left(d_0,\operatorname{dist}(p_i,p_j)s_c\right),\\
\mathrm{PL}(d_m)&=L_0+10\eta\log_{10}(d_m/d_0),\\
P_{\mathrm{rx}}&=P_{\mathrm{tx}}-\mathrm{PL}(d_m)+10\log_{10}h,
\end{align}
where $L_0=40$ dB, $\eta=3.0$, $d_0=1$ m, $s_c=1$ m/cell, $P_{\mathrm{tx}}=30$ dBm, and $h\sim\operatorname{Exp}(1)$. Delivery occurs when $P_{\mathrm{rx}}\geq P_{\mathrm{sens}}$~\cite{1996rappaport}. Peer-state and allocator-specific messages pass through the active loss model. Collision-intent messages retain communication delay but bypass loss.

\section{Algorithms}
\label{sec:algorithms}

\subsection{Common Allocator Interface and Cost Model}

All methods are evaluated under the same environment, movement model, local mission state, communication layer, and termination condition. Each implementation acts only as a task allocator. At a planning event, it returns a target cell, while the shared robot shell handles A* path planning, movement, target completion, local-state updates, collision avoidance, communication delivery, and metric logging. Physical execution assumptions therefore remain constant across algorithms.

The common allocation edge cost is $C(a,x)=d_1(a,x)$, where $a$ is the robot's current cell or a route-reference cell and $x\in T$ is an active target. CBAA and ACBBA use the negative of this cost as a bid. HIPC applies the corresponding route score during local team planning. PI uses the same distance in route significance, DMCHBA places it in the Hungarian cost matrix, and DGA includes it in team-plan fitness.

Algorithms that require allocation communication use their native protocol messages. Collision-intent messages belong to the shared safety layer. They retain communication delay and bypass loss, but they are not available to an allocator as ordinary peer state. Table~\ref{tab:algorithms} summarizes the planning and repair mechanisms described below.

\begin{table*}[!t]
\centering
\caption{Benchmark allocation and communication mechanisms.}
\label{tab:algorithms}
\scriptsize
\setlength{\tabcolsep}{3.3pt}
\renewcommand{\arraystretch}{1.10}
\begin{tabularx}{\textwidth}{@{}p{0.09\textwidth}p{0.17\textwidth}p{0.29\textwidth}Y@{}}
\toprule
\textbf{Method} & \textbf{Planning scope} & \textbf{Local allocation mechanism} & \textbf{Allocator communication and repair}\\
\midrule
CBAA~\cite{2009choi} & One current target & Highest locally feasible distance-based bid & Delta winner--bid records, deterministic tie breaking, and rebidding after ownership loss\\
ACBBA~\cite{2010johnson} & Ordered bundle & Marginal route-insertion bids & Timestamped delta deconfliction and release of a lost target with its dependent suffix\\
PI~\cite{2016zhao} & Ordered path & Marginal route significance, with lower significance preferred & Owner, significance, and path records with local removal and rescoring after ownership loss\\
HIPC~\cite{2016johnson} & Local team plan & Greedy planning over the local robot and predictable peers & Path claims, prediction checking, stale-claim cleanup, and neighborhood pruning after repeated mismatch\\
DMCHBA~\cite{2024samiei} & Clone-expanded assignment & Hungarian matching followed by local route ordering & No allocator-specific bid or assignment messages, with allocation reconstructed from locally available team state\\
DGA~\cite{2020patel} & Complete team plans & Population search with selection, crossover, mutation, repair, and elitism & Changed owner-route information is exchanged and valid received plans enter the local population\\
\bottomrule
\end{tabularx}
\end{table*}

\subsection{CBAA}

The Consensus-Based Auction Algorithm (CBAA)~\cite{2009choi} is implemented as a single-task allocator with auction and consensus phases. At each planning event, each robot computes a distance-based bid for every valid target and selects the best target for which its bid exceeds the locally stored winning value. The robot records the target, bid, and winner in its local table. Changed entries are transmitted as deltas rather than as a full table, and a robot may forward a valid update learned from a peer even when it is not the reported winner.

Received entries replace local records when they contain a higher bid or win the deterministic tie break. Release messages clear only matching stale ownership and cannot erase a stronger competing claim. When consensus shows that a robot no longer owns its current target, the assignment is cleared and the robot may bid again at a later planning event.

\subsection{ACBBA}

The Asynchronous Consensus-Based Bundle Algorithm (ACBBA)~\cite{2010johnson} extends bid consensus to ordered multi-task bundles. Each robot builds a route up to planning depth $B$ by evaluating every insertion position for each candidate target. The insertion with the highest marginal bid is selected. The first target in the resulting path becomes the immediate assignment, while later targets represent planned future work.

Each robot maintains a local winner, bid, and timestamp for every target. Changed records are communicated as deltas and processed using the asynchronous winner, bid, and timestamp rules. When a robot is outbid on a target in its bundle, that target and all later elements are released because their marginal values depend on the retained prefix. This suffix-release rule distinguishes ACBBA repair from the local rescoring used by PI.

\subsection{PI}

The Performance Impact (PI) algorithm~\cite{2016zhao} is implemented as a significance-based multi-task allocator. Each robot maintains an ordered path of up to planning depth $B$. A candidate target is evaluated at every insertion position, and the least costly insertion defines its marginal impact. The robot then selects the target whose insertion provides the greatest improvement relative to its locally stored significance table. Lower significance is preferred because it represents a smaller contribution to route cost.

Robots communicate ownership, significance, timestamps, and current path membership for the targets they retain. Newer reports from the same owner replace stale records, while competing claims are resolved by significance and deterministic tie breaking. Losing a target does not invalidate the remainder of the path. PI removes only that target and recomputes the significance of the surviving route.

\subsection{HIPC}

The Hybrid Information and Plan Consensus (HIPC) algorithm~\cite{2016johnson} combines implicit local-team planning with plan consensus. Each robot forms a planning neighborhood from itself and peers whose behavior can be estimated from received state. It then applies a greedy team-level allocation over that neighborhood, assigning up to planning depth $B$ targets to each modeled robot. Only the portion assigned to the local robot is retained, and its first target becomes the immediate assignment.

The local state records a winner, bid, and timestamp for each target. Path claims include current membership so receivers can remove stale claims from the same sender. HIPC also compares received peer paths with its local predictions. Peers that repeatedly depart from those predictions are temporarily excluded from local team planning, although their communicated claims remain respected. Losing an earlier target releases the dependent suffix of the retained path.

\subsection{DMCHBA}

The Distributed Matching-by-Clone Hungarian-Based Algorithm (DMCHBA)~\cite{2024samiei} is implemented as an implicit-coordination allocator that extends the one-to-one DHBA formulation~\cite{2017ismail}. When reassignment is triggered, each robot creates enough clones of every known agent to cover the active target set, adds pseudotasks when needed, and solves the resulting square minimum-cost assignment with the Hungarian method. During online CV execution, each robot follows up to three targets from its current planned route before triggering a new allocation.

DMCHBA sends no allocator-specific bid, bundle, or winner-table messages. Each robot reconstructs the assignment from locally available shared state. The targets assigned to its own clones are ordered greedily into an executable route, and the first target becomes the immediate assignment. 

\subsection{DGA}

The Decentralized Genetic Algorithm (DGA)~\cite{2020patel} is implemented as a population-based allocator in which each robot maintains complete ordered team plans. When replanning is triggered, the robot repairs current and received plans, adds greedy and randomized seeds, and evolves the population through elitism, tournament selection, recombination, mutation, and repair.

Candidate plans are ranked according to the evaluated route objective. MinMax comparisons use $J_{\max}+0.05J_{\Sigma}$, while MinSum comparisons use $J_{\Sigma}+0.05J_{\max}$, retaining the alternate route objective as a lower-weight secondary term. After evolution, the robot executes the route assigned to itself in its best local plan. Changed owner-route information is communicated through deltas rather than full populations. Receivers reconstruct valid routes, repair them against local mission state, and retain improved plans for later evolution. Unless otherwise stated, pre-allocation reliability and computation and the primary asynchronous timing comparison use the MinMax score, while MinSum route-performance results use the MinSum score.

DGA uses 25 generations per triggered update, selected from the accompanying iteration-sensitivity study available at \url{https://github.com/jlott22/CV_MRTA_Benchmark/DGA_tuning}. During online CV execution, each agent retains up to the next three targets from its selected team plan, while improved received solutions or execution events can trigger earlier replanning. The remaining GA settings use a population of 30, crossover probability 0.70, mutation probability 0.30, tournament size three, and two elite individuals, consistent with prior GA parameter studies~\cite{1986grefenstette,2023gnanapragasam}.

\subsection{Planning-Depth Selection}

For ACBBA, HIPC, and PI, planning depth was selected independently for each algorithm $a$ and route objective $o\in\{\Sigma,\max\}$ using the companion planning-depth study~\cite{lott2026}:
\begin{equation}
\begin{aligned}
B_{a,o}^{*}
&=\arg\min_{B\in\{2,3,5,8,12\}}
\frac{1}{2N_{\mathrm{tune}}}
\sum_{i=1}^{N_{\mathrm{tune}}}\\[-1mm]
&\quad\left[Y_{a,o,i}^{\mathrm{ideal}}(B)
+Y_{a,o,i}^{\mathrm{Bern}(0.25)}(B)\right].
\end{aligned}
\label{eq:bundle_selection}
\end{equation}
Here $N_{\mathrm{tune}}=300$. The observed objective is $Y_{a,\Sigma,i}=\sum_{r\in R}s_{r,i}$ for MinSum and $Y_{a,\max,i}=\max_{r\in R}s_{r,i}$ for MinMax, where $s_{r,i}$ is the travel completed by robot $r$ in tuning trial $i$. Ideal and Bernoulli conditions receive equal weight, and ties are resolved toward the smaller value. The candidate range begins at $B=2$ so all three methods retain multi-task planning. The selected MinSum depths are 2, 8, and 5 for ACBBA, HIPC, and PI, respectively, while all three use depth 2 for MinMax. These are parameter settings of the same algorithms, and each remains fixed within the corresponding objective across communication and scale conditions. Depth selection uses only ideal delivery and Bernoulli $p_{\mathrm{drop}}=0.25$; Gilbert--Elliott and Rayleigh conditions do not enter Eq.~\eqref{eq:bundle_selection} and are not separately retuned. Results under those communication families therefore evaluate transfer of the fixed selected depths beyond the channel conditions used for selection rather than channel-specific tuning. CBAA is intrinsically single-task, while DMCHBA and DGA do not use this parameter.

\section{Experimental Methodology}

\subsection{Experimental Campaigns}

Table~\ref{tab:campaigns} organizes four campaign groups. The asynchronous core benchmark measures realized mission performance and coordination under communication degradation. The standardized pre-allocation study measures viable and stable outcomes, full-protocol computation, and achieved route quality. A separate asynchronous timing campaign records computation accumulated during execution, while complementary scale-sensitivity studies test whether the objective rankings persist across changes in workspace geometry, robot density, and target load.

\begin{table*}[!t]
\centering
\caption{Experimental campaigns.}
\label{tab:campaigns}
\scriptsize
\setlength{\tabcolsep}{4pt}
\renewcommand{\arraystretch}{1.10}
\begin{tabularx}{\textwidth}{@{}p{0.19\textwidth}p{0.16\textwidth}p{0.22\textwidth}p{0.16\textwidth}Y@{}}
\toprule
\textbf{Campaign} & \textbf{Algorithms} & \textbf{Conditions} & \textbf{Sampling} & \textbf{Purpose}\\
\midrule
Asynchronous core & Six & Ten targets under ideal delivery and 24 degraded communication conditions & 500 per condition, 112,500 unique simulations & Realized travel, communication, coordination, and degradation response\\
Synchronous pre-allocation & Five multi-task methods & 5, 10, 25, and 50 targets under selected ideal and degraded treatments & 50, 100, 50, and 25 per condition, 4,500 trials & Viable and stable allocation, $\Cterm$, and achieved route quality\\
Asynchronous computation & Six & 5, 10, 25, and 50 targets under ideal delivery & 1,350 primary records plus 675 objective-specific checks & Execution-integrated computation and allocator-call frequency\\
Scale sensitivity & Six & Grid-size/robot-density sweep and fixed-grid target-load/robot-count sweep & 14,400 grid/density simulations plus 3,200 target/team simulations & MinMax and MinSum ranking continuity across scale \\
\bottomrule
\end{tabularx}
\end{table*}

The core benchmark uses one ideal condition and eight nominal severity levels for each degraded communication model,
\begin{equation}
q\in\{5,10,20,30,40,50,60,70\}.
\end{equation}
For Bernoulli and Gilbert--Elliott loss, $p_{\mathrm{drop}}=q/100$ is the configured stationary drop fraction. The Rayleigh fading receiver thresholds are
\begin{multline}
P_{\mathrm{sens}}\in\{-59.40,-56.04,-52.15,-49.17,\\
-46.04,-42.16,-37.79,-32.58\}\ \mathrm{dBm}.
\end{multline}
These thresholds define an ordered degradation trajectory and are not treated as realized-loss matches to the other communication models. The 500 nominal ten-target layouts were sampled sequentially from a reproducible random-number stream initialized with seed 20260630. Each trial execution used a distinct deterministic simulator seed derived from its trial ID, with the same layout and trial seed reused across algorithms and communication conditions to preserve the paired design. For each trial, ten distinct target cells were sampled uniformly without replacement from the $19\times19$ grid after excluding the four robot starting cells. No minimum inter-target spacing or clustering constraint was imposed, and the nominal workspace contains no obstacles; therefore adjacent, boundary, and spatially clustered targets may occur naturally. Each trial identifier reuses the resulting target layout and robot starts across algorithms and communication conditions.

One algorithm--setting combination contributes $500\times25=12{,}500$ runs. CBAA, ACBBA, and DMCHBA use the same implemented setting for both route objectives and therefore contribute $3\times12{,}500=37{,}500$ shared records. HIPC, PI, and DGA are each evaluated separately under their MinMax and MinSum settings, contributing $3\times2\times12{,}500=75{,}000$ objective-specific records. The core campaign therefore contains 112,500 unique simulations. For either route objective, the six-algorithm comparison combines the 37,500 shared records with the corresponding 37,500 objective-specific records for HIPC, PI, and DGA, yielding 75,000 records. The underlying cost functions and all other benchmark conditions remain standardized across methods and objectives.

The scale-sensitivity study contains two complementary components. The grid/density campaign uses grid side lengths $g\in\{14,19,25,34\}$ and nominal cells-per-robot levels $\delta\in\{50,85,140,220\}$, with $N_R=\lceil g^2/\delta\rceil$. Every condition retains ten targets and is evaluated under ideal delivery and Bernoulli $p_{\mathrm{drop}}=0.25$. Of 14,400 unique simulations, 14,383 completed. Continuous MinMax and MinSum comparisons retain 1,592 and 1,589 complete six-algorithm condition--trial blocks, respectively, from the 1,600 designed blocks. The target-load/team-size campaign fixes the workspace at $19\times19$ under ideal delivery and varies target count over $M\in\{25,50,75,100\}$ and robot count over $N_R\in\{6,8,12,16\}$. Each combination uses 25 paired target layouts shared across algorithms. HIPC and PI use the planning depths selected for the corresponding route objective, while CBAA, ACBBA, DMCHBA, and DGA provide common records for both objective views, producing 3,200 planned unique simulations. The nominal ten-target, four-robot reference is taken from the existing core benchmark rather than rerun.

\subsection{Mission and Coordination Metrics}

Mission movement is recorded as the number $s_i$ of completed grid transitions made by robot $r_i$ at termination. Realized MinMax and MinSum movement are
\begin{equation}
\Smax=\max_{r_i\in R}s_i, \qquad
\Ssum=\sum_{r_i\in R}s_i.
\label{eq:realized}
\end{equation}
All robots use the same translation, turning, communication, and event-timing model. Under this common timing model, $\Smax$ provides a discrete proxy for mission makespan, while $\Ssum$ measures aggregate locomotion effort. The MinMax and MinSum results use the algorithm settings associated with the corresponding route objective. Associated communication and coordination outcomes are calculated from the same runs.

Communication burden is represented by total sender publications $\Msum$ and publication intensity,
\begin{equation}
\Mstep=\frac{\Msum}{\Ssum}.
\end{equation}
A publication is counted once at the sender regardless of the number of receivers and includes shared-core and allocator-specific traffic. $\Msum$ measures total publication events, while $\Mstep$ measures the publication demand associated with one team movement step.

Redundant work is counted whenever a robot reaches a target after the team's first physical completion of that target. Within-team workload concentration is described using target-completion Gini, calculated from the number of first target completions credited to each robot. A Gini coefficient of zero represents equal contributions, while larger values represent greater concentration among team members.

To summarize degradation trajectories, the analysis uses Paired Relative Degradation Slope (PRDS) and Paired Relative Degradation Area (PRDA). Let $\mathcal{Q}=\{0,5,10,20,30,40,50,60,70\}$ denote the nominal severity levels, where $q=0$ denotes ideal delivery. For algorithm $a$, paired trial $t$, metric $m$, and severity $q$, define
\begin{equation}
Z_{a,t}(q)=\log\left(Y_{a,t}(q)+c_m\right),
\end{equation}
where $c_m=0$ for strictly positive outcomes, including $\Smax$ and $\Ssum$, and $c_m=1$ for outcomes that may be zero. The within-algorithm trajectory is fitted as
\begin{equation}
Z_{a,t}(q)=\alpha_{a,t}+\beta_{a,t}q+\varepsilon_{a,t}(q),
\end{equation}
and
\begin{equation}
\mathrm{PRDS}_{a,t}=100\left[\exp\left(\widehat{\beta}_{a,t}\right)-1\right].
\end{equation}
PRDS estimates the multiplicative percentage change associated with a one-percentage-point increase in nominal severity along an algorithm's own trajectory. Lower PRDS indicates less within-algorithm degradation.

Whereas PRDS follows an algorithm's own trajectory, PRDA measures how its degradation differs from the contemporaneous six-algorithm field. At each severity,
\begin{equation}
\overline{Z}_{t}(q)=\frac{1}{|\mathcal{A}|}
\sum_{b\in\mathcal{A}}Z_{b,t}(q),
\qquad |\mathcal{A}|=6,
\end{equation}
and the ideal-centered relative trajectory is
\begin{equation}
D_{a,t}(q)=
\left[Z_{a,t}(q)-\overline{Z}_{t}(q)\right]
-
\left[Z_{a,t}(0)-\overline{Z}_{t}(0)\right].
\end{equation}
PRDA is calculated by trapezoidal integration,
\begin{equation}
\mathrm{PRDA}_{a,t}=
\frac{100}{70}
\operatorname{Trapz}_{q\in\mathcal{Q}}D_{a,t}(q).
\end{equation}
Negative PRDA indicates that the algorithm improves its relative position against the six-method field as degradation increases, while positive PRDA indicates greater relative degradation. Both summaries require a complete paired trajectory containing ideal delivery and all eight impaired levels for all six algorithms, producing $n=500$ eligible trajectories for each communication model. Rayleigh values follow the ordered receiver-sensitivity schedule and are interpreted within that communication family.

\subsection{Standardized Pre-Allocation Protocol}
\label{sec:prealloc}

In the standardized pre-allocation study, ACBBA, PI, HIPC, DMCHBA, and DGA receive a common complete static routing problem. CBAA is excluded because its single-task state does not represent a simultaneous complete multi-target allocation. All target coordinates and four robot starts are available at time zero, allocator-local state begins empty, and the robots remain fixed. ACBBA, PI, and HIPC may represent up to the active target count, DMCHBA exposes its complete clone-based matching, and DGA represents complete team plans. The online planning-depth limits are nonbinding in this campaign and therefore do not affect the pre-allocation results.

Each allocation attempt proceeds in synchronized communication rounds. Every robot first updates from information available at the start of the round and generates its native messages. The outbound batch is then frozen, directed deliveries are evaluated, surviving messages are processed, and a read-only observer composes the executable team allocation from the robots' local routes. Communication loss applies to allocator-specific coordination messages. DMCHBA reconstructs its deterministic assignment from the common static inputs and therefore generates no allocator-specific message in this protocol. DGA uses versioned complete-plan frames. A received plan is adopted only after all fragments have been reconstructed, target coverage has been validated, and fitness has been recomputed locally.

Viability is reached when every active target appears exactly once across complete, conflict-free, executable routes. A viable allocation is \emph{stable} when the same ordered allocation appears at three consecutive end-of-round checks and then remains unchanged for ten additional rounds. Any intervening change restarts the stability count. Trials stop when stability is established or when the protocol reaches round 500.

Full-protocol computation is represented by $\Cterm$, the cumulative native allocator wall computation over the complete static deliberation protocol. For a stable trial, accumulation ends when the validation interval is complete. Otherwise, it ends at the round-500 cap. It includes native local allocator updates, allocator-message generation, and delivered-message processing. The scheduler, observer, artificial communication delay, and file input/output are excluded. Accordingly, $\Cterm$ measures the full protocol's computation budget and includes every planned attempt in its condition.

The primary ten-target study contains 100 paired scenarios under ideal delivery, Bernoulli $p_{\mathrm{drop}}=0.25$, a frozen Gilbert--Elliott condition with $p_{GG}=0.95$ and $p_{BB}=0.85$, Rayleigh fading at $P_{\mathrm{sens}}=-50.66$ dBm, and Bernoulli $p_{\mathrm{drop}}=0.50$. Five- and 25-target sensitivity conditions use 50 scenarios under ideal and Bernoulli $p_{\mathrm{drop}}=0.25$, while the 50-target stress condition uses 25. All 4,500 trials ran serially with one measured worker and one numerical thread.

Route quality is calculated from the open Manhattan routes of stable allocations. The five-method omnibus tests and displayed route-quality summaries use the 68 ten-target ideal scenarios in which every method produced a stable allocation and finite route costs. This common subset preserves direct five-method comparability. Viable and stable counts and $\Cterm$ summaries use every planned attempt.

\subsection{Asynchronous Mission Computation}

During the asynchronous timing campaign, every allocator invocation generated by complete ideal-communication CV execution is measured. The primary MinMax comparison contains 50 paired missions at five targets, 100 at ten targets, 50 at 25 targets, and 25 at 50 targets, producing 1,350 algorithm--mission records. An additional 675 records evaluate HIPC, PI, and DGA under the MinSum objective. Both sets use the same 12-worker batch protocol. Timing uses a monotonic high-resolution clock around the allocator entry point and does not advance simulated time or change event ordering.

For robot $i$ in trial $t$, let $c_{i,t,k}$ be the duration of its $k$th allocator call and $n_{i,t}$ its number of calls. The reported trial-level outcomes are
\begin{align}
C^{\mathrm{mission}}_{\Sigma,t}
&=\sum_{i\in R}\sum_{k=1}^{n_{i,t}}c_{i,t,k},\\
N_{C,t}
&=\sum_{i\in R}n_{i,t}.
\end{align}
$C^{\mathrm{mission}}_{\Sigma,t}$ measures total processor work accumulated across the team, while $N_{C,t}$ measures allocator-invocation frequency. Their joint interpretation separates expensive individual decisions from inexpensive decisions that are repeated many times. All planned timing missions completed.

All wall-clock computation measurements were collected on a Windows 11 laptop with an Intel Core Ultra 7 155H processor (16 cores, 22 threads) and 16~GB RAM, using Python 3.13.14 with NumPy 2.3.3 and pandas 3.0.3. The pre-allocation $\Cterm$ study used one worker and one numerical thread, whereas the asynchronous timing study used 12 worker processes with one numerical thread each; accordingly, reported runtimes characterize the tested implementations and hardware rather than language-independent algorithmic complexity.

The source code, experiment configurations, and analysis scripts used in this study are available at
\url{https://github.com/jlott22/CV_MRTA_Benchmark}.
Detailed allocator communication tables are provided at
\url{https://github.com/jlott22/CV_MRTA_Benchmark/communication_tables},
and the DGA iteration-sensitivity study, including results and figures, is available at
\url{https://github.com/jlott22/CV_MRTA_Benchmark/DGA_tuning}.

\subsection{Statistical Analysis}

Throughout the analysis, scenario is the experimental replicate. For each continuous core outcome and communication condition, a Friedman test evaluates the overall six-algorithm difference~\cite{1937friedman}, Kendall's $W$ reports the omnibus effect size~\cite{1939kendall}, and paired Wilcoxon signed-rank tests identify pairwise differences~\cite{1945wilcoxon}. Holm correction controls familywise error within each prespecified outcome--condition family~\cite{1979holm}. Rank-biserial correlation describes effect direction~\cite{1956cureton}, while paired scenario-bootstrap confidence intervals quantify uncertainty~\cite{1979efron}. Counts of condition leads and corrected pairwise wins are reported as cross-condition summaries. For each route objective, every statistical family contains one result per algorithm under the corresponding objective settings.

Each complete trial-level degradation trajectory yields one PRDS value, and algorithm-level intervals use two-sided 95\% Student $t$ intervals. PRDA intervals use paired-trial percentile-bootstrap intervals from 10,000 resamples.

For pre-allocation, viable and stable counts are reported over all planned trials. Viable counts are descriptive. Paired stable/not-stable outcomes are compared with McNemar tests~\cite{1947mcnemar}, with Holm correction within each prespecified condition family. The five-method route-quality omnibus tests and displayed summaries use the 68 common stable ten-target ideal scenarios. Pairwise route-quality tests use the largest matched subset stable for both methods in each contrast, producing sample sizes from 68 to 100, and use Holm-adjusted paired Wilcoxon tests with paired bootstrap intervals.

All-attempt $\Cterm$ summaries include every planned trial. For each algorithm, degraded-versus-ideal differences in the ten-target study are evaluated with paired two-sided Wilcoxon tests. Holm correction controls familywise error across the four communication treatments. Cross-algorithm $\Cterm$ comparisons are descriptive because the endpoint combines stability-validation work with round-cap work for attempts that do not become stable.

Target-load comparisons, static-to-asynchronous computation rankings, and the scale-sensitivity studies are interpreted descriptively. Different target loads use independently generated scenario sets, while the static and asynchronous studies measure different computational quantities.

\section{Results and Discussion}

\subsection{Asynchronous Mission Performance}

\begin{figure*}[!t]
\centering
\includegraphics[width=0.90\textwidth]{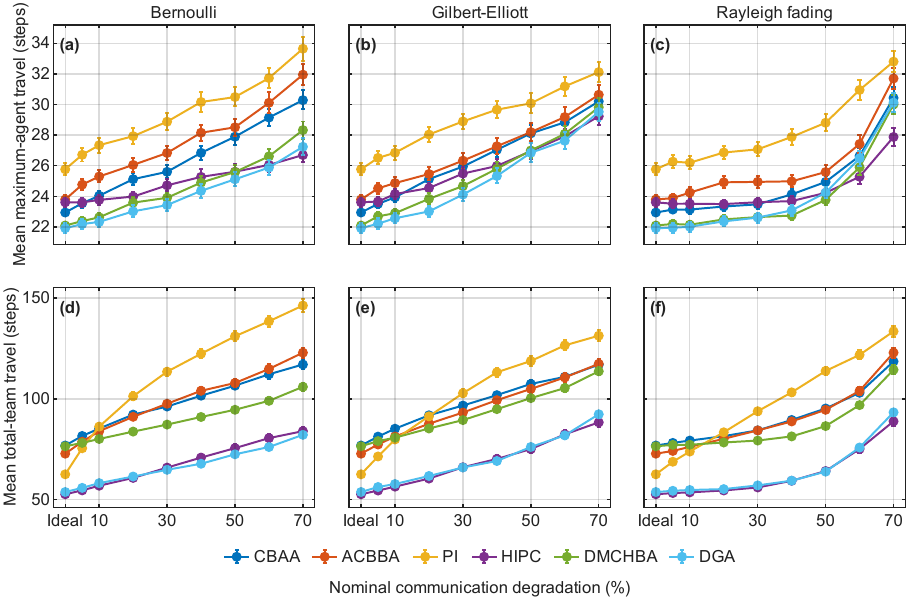}
\caption{Collaborative Visit travel across ideal and degraded communication. Panels (a)--(c) show realized MinMax travel $\Smax$, and panels (d)--(f) show realized MinSum travel $\Ssum$. Each objective uses the corresponding algorithm settings. Points are paired-scenario means with 95\% bootstrap confidence intervals.}
\label{fig:core}
\end{figure*}

Across the 24 impaired conditions, DGA produced the lowest unweighted mean $\Smax$ at 24.49 steps, followed by DMCHBA at 24.78 and HIPC at 25.12. As shown in Fig.~\ref{fig:core}(a)--(c), DGA led 18 conditions, HIPC led four, and DMCHBA led two. The paired DGA--DMCHBA comparison favored DGA in 12 conditions after Holm correction. No condition favored DMCHBA. Their across-condition means nevertheless differed by only 0.29 steps, identifying DGA and DMCHBA as the leading MinMax pair. DGA had the stronger condition-level record, while DMCHBA remained close in realized travel. HIPC ranked third overall and entered the top two in eight conditions, with its separation narrowing at the highest loss levels. Table~\ref{tab:core_summary} summarizes the impaired-condition MinMax and publication-intensity results.

The ordering changed for total-team travel. Figure~\ref{fig:core}(d)--(f) shows HIPC and DGA forming a closely matched leading pair across the three communication models. Their impaired-condition mean $\Ssum$ values were 66.95 and 67.22 steps, respectively, followed by DMCHBA at 89.99. ACBBA and CBAA followed at 95.73 and 96.42 steps, while PI averaged 105.92. Under ideal delivery, DGA ranked second at 53.75 steps and PI ranked third at 62.54. HIPC led 15 impaired condition means and DGA led nine, with both occupying the top two in every impaired condition. The paired HIPC--DGA comparisons favored HIPC in 13 conditions and DGA in two after within-condition Holm correction, while nine did not reach significance. DGA and DMCHBA therefore form the leading MinMax pair, while HIPC and DGA share the leading MinSum region.

\begin{table}[htbp]
\centering
\caption{Impaired-condition asynchronous MinMax summary.}
\label{tab:core_summary}
\scriptsize
\setlength{\tabcolsep}{3.2pt}
\begin{tabular}{lrrrr}
\toprule
\textbf{Method} & $\boldsymbol{\Smax}$ & \textbf{Leads} & \textbf{Top two} & $\boldsymbol{\Mstep}$\\
\midrule
CBAA   & 26.02 & 0  & 0  & 5.19\\
ACBBA  & 26.91 & 0  & 0  & 3.71\\
PI     & 29.05 & 0  & 0  & 2.48\\
HIPC   & 25.12 & 4  & 8  & 2.61\\
DMCHBA & 24.78 & 2  & 19 & 2.08\\
DGA    & 24.49 & 18 & 21 & 3.38\\
\bottomrule
\end{tabular}

\vspace{4pt}
\parbox{\linewidth}{\footnotesize
\emph{Note:} Values use the MinMax settings. $\Smax$ and $\Mstep$ are unweighted means of the 24 impaired-condition means. ``Leads'' and ``Top two'' count conditions in which a method had the lowest or one of the two lowest mean $\Smax$.
}
\end{table}

\subsection{Communication and Coordination Behavior}

\subsubsection{MinMax Travel and Publication Intensity}

\begin{figure}[htbp]
\centering
\includegraphics[width=\columnwidth]{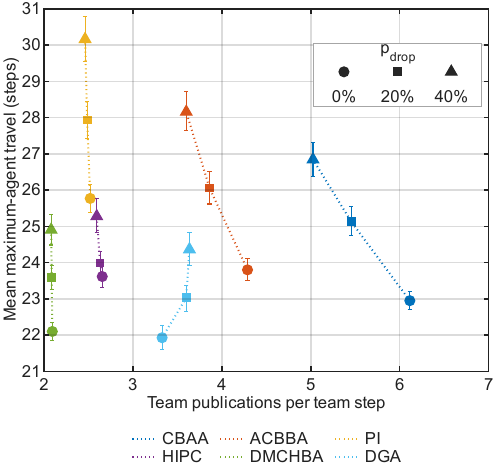}
\caption{Realized MinMax travel versus publication intensity under ideal delivery and Bernoulli loss at $p_{\mathrm{drop}}=0.20$ and $0.40$. Lower values on both axes are preferred. Error bars are 95\% paired-scenario bootstrap confidence intervals.}
\label{fig:comm_message_tradeoff}
\end{figure}

Figure~\ref{fig:comm_message_tradeoff} traces the tradeoff between MinMax travel and publication intensity under ideal delivery and Bernoulli loss at $p_{\mathrm{drop}}=0.20$ and $0.40$. DGA retained the lowest mean $\Smax$ at all three levels, increasing from 21.93 to 24.36 steps, while its publication intensity increased from 3.33 to 3.64 publications per team step. DMCHBA remained close in travel performance, increasing from 22.10 to 24.90 steps, while maintaining the lowest and nearly constant publication intensity at approximately 2.09 publications per team step. Together, DGA and DMCHBA form the principal frontier over the displayed range.

At $p_{\mathrm{drop}}=0.40$, HIPC occupied a nearby low-publication region at 25.28 steps and 2.59 publications per team step, although DMCHBA remained lower on both displayed outcomes. PI communicated at a similarly low intensity but incurred substantially greater MinMax travel, while CBAA and ACBBA required more publications.

\subsubsection{Robustness Under Message Loss}

\begin{figure}[htbp]
\centering
\includegraphics[width=\columnwidth]{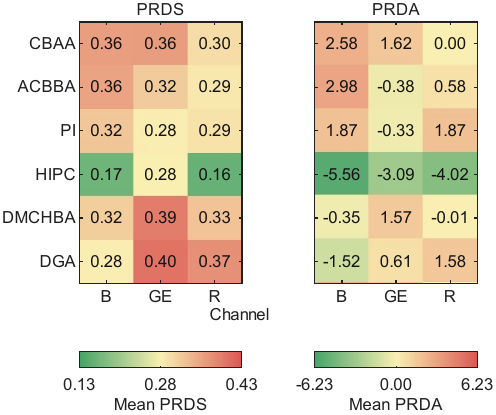}
\caption{MinMax communication-degradation summaries. Columns B, GE, and R denote Bernoulli, Gilbert--Elliott, and Rayleigh fading communication. Lower PRDS indicates less within-algorithm degradation. Negative PRDA indicates improving position relative to the six-algorithm field.}
\label{fig:heatmap}
\end{figure}

When robustness was measured relative to each method's ideal baseline, HIPC exhibited the most favorable response in realized MinMax. It had the lowest or tied-lowest PRDS across all three communication models and the most favorable PRDA under each model, indicating that its relative position improved as communication worsened. PI also followed comparatively flat MinMax trajectories, but retained the highest absolute $\Smax$ in every communication condition. DGA and DMCHBA degraded more rapidly relative to their ideal baselines while maintaining the strongest absolute MinMax performance over most of each sweep. Figure~\ref{fig:heatmap} therefore distinguishes absolute performance under impaired communication from sensitivity relative to an allocator's ideal operating point. HIPC narrowed its MinMax gap most consistently at severe loss, while the relative positions of DGA and DMCHBA varied across the independent, bursty, and distance-dependent models.

For MinSum, relative robustness differed from absolute route performance. DMCHBA had the lowest PRDS and most favorable PRDA under all three communication families, indicating the smallest degradation relative to both its own ideal baseline and the contemporaneous algorithm field. DGA had the second-most favorable PRDA in all three families, although its within-algorithm degradation was less favorable under Gilbert--Elliott and Rayleigh fading. HIPC remained one of the two leading methods in absolute MinSum travel, but its relative degradation measures were less favorable than those of DMCHBA and DGA. PI showed the strongest degradation by both measures, with PRDS above 1\% per severity point in all three families and strongly positive PRDA. As with MinMax, these results distinguish absolute route quality from robustness relative to an algorithm's nominal performance.

\subsubsection{Coordination Efficiency}

Redundant target visits and workload concentration help explain the route-performance differences observed in the MinMax and MinSum results. Redundant visits add travel without contributing a new target completion and can therefore penalize both objectives. However, workload concentration affects them differently. Distributing work more evenly can limit the longest individual route and support MinMax, whereas MinSum can benefit when spatially compatible targets are concentrated on fewer robots and unnecessary cross-team travel is avoided.

The leading MinSum algorithms also produced the fewest redundant visits, although their ordering differed. DGA had the lowest impaired mean at 1.150 duplicate target visits per mission, followed by HIPC at 1.276, despite HIPC's slightly lower aggregate travel. PI's high mean of 3.412 was consistent with its weaker impaired MinSum performance. In MinMax, DGA achieved the lowest redundant-target-visit mean at 1.353, matching its leading performance in that objective. HIPC was nearly identical at 1.354 and outperformed DMCHBA at 1.713 despite DMCHBA's lower MinMax travel. Redundant work therefore tracks route performance closely in some cases, but does not by itself explain the objective ordering.

Under the MinSum operating point, HIPC had the highest target-completion Gini at 0.493, while DGA also exhibited concentrated workload at 0.415. HIPC's Gini fell to 0.267 under the MinMax objective, consistent with results in \cite{lott2026} showing that MinMax and MinSum favor different workload distributions. DGA's MinMax Gini was 0.246, leaving DGA and DMCHBA with the two lowest coefficients in that comparison. Together, these results associate the leading MinSum solutions with concentrated, low-redundancy allocation, whereas the MinMax leaders distribute completed work more evenly.

\subsection{Static Pre-Allocation Reliability, Route Quality, and Computation}
\label{subsec:preallocation_results}

Static pre-allocation reveals whether each multi-task method produces a viable executable allocation, whether that allocation remains stable, what route quality it achieves, and how much native computation the complete protocol consumes. Because the planning-depth limit is nonbinding in this protocol, these results are independent of the online depth selection. Route quality is evaluated only for stable allocations. Across the 4,500 trials, 1,447 attempts reached the round-500 cap, and 1,430 ended with an incomplete or conflicted allocation, so their terminal route costs do not represent comparable solutions. These attempts remain part of the viable/stable and $\Cterm$ results.

\subsubsection{Viable and Stable Allocation Outcomes}

Under ideal delivery, all five methods were generally able to produce complete allocations. HIPC, DMCHBA, and DGA were viable and stable in every trial, ACBBA was stable in 98 of 100, and PI showed the largest separation between reaching a viable allocation and maintaining it through validation, with 97 viable but only 70 stable outcomes.

Communication degradation separated the methods more clearly, with HIPC and DMCHBA remaining viable and stable in every ten-target treatment, while ACBBA, PI, and DGA deteriorated as loss increased. ACBBA and PI often reached viable allocations that did not remain stable, whereas DGA more often failed before a complete allocation was established. At Bernoulli $p_{\mathrm{drop}}=0.50$, for example, HIPC and DMCHBA remained at 100/100 viable/stable outcomes, compared with 35/9 for ACBBA, 18/3 for PI, and 5/4 for DGA in the primary comparison.

\begin{figure}[!t]
\centering
\includegraphics[width=\columnwidth]{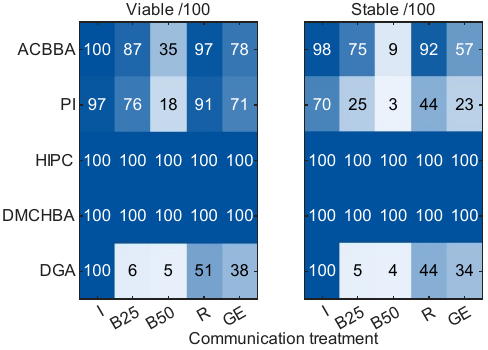}
\caption{Ten-target pre-allocation reliability across five communication treatments. Panels (a) and (b) report viable and stable allocations, respectively, out of 100 paired attempts.}
\label{fig:prealloc_viable_stable}
\end{figure}

\subsubsection{Route Quality and Full-Protocol Computation}

For the primary ten-target ideal condition, Table~\ref{tab:prealloc_summary} reports reliability, route quality, and full-protocol computation. The route-quality comparison uses the 68 scenarios in which all five methods produced stable allocations. These summaries are conditional on joint stability and need not represent the full scenario set. Pairwise tests use the largest matched subset stable for each pair, with 68–100 scenarios, while reliability and full-protocol computation retain all planned attempts. DGA achieved the lowest median MinMax cost at 21.0 steps, followed by DMCHBA at 22.5. ACBBA and HIPC each required 40.0, while PI required 48.0. The overall MinMax difference was strong, with $\chi^2=220.63$, $p=1.37\times10^{-46}$, and $W=0.811$.

On the MinSum objective, DGA achieved the lowest median at 50.0 steps, followed by HIPC at 51.5, PI at 56.0, ACBBA at 61.5, and DMCHBA at 70.0. The overall difference was strong, with $\chi^2=220.22$, $p=1.68\times10^{-46}$, and $W=0.810$. The paired DGA--HIPC comparison over the 100 scenarios stable for both methods also favored DGA after Holm correction ($p=9.54\times10^{-8}$). Unlike the narrow HIPC lead in the asynchronous core aggregate, the ten-target static comparison therefore favored DGA on MinSum.

\begin{table}[!t]
\centering
\caption{Ten-target ideal pre-allocation summary.}
\label{tab:prealloc_summary}
\scriptsize
\setlength{\tabcolsep}{6.0pt}
\renewcommand{\arraystretch}{1.10}
\begin{tabular}{lrrrrr}
\toprule
\textbf{Method} &
\makecell{\textbf{Viable}\\\textbf{/100}} &
\makecell{\textbf{Stable}\\\textbf{/100}} &
\makecell{$\boldsymbol{\Cterm}$\\\textbf{(ms)}} &
\textbf{MinMax} &
\textbf{MinSum} \\
\midrule
ACBBA  & 100 & 98  & 86.62   & 40.0 & 61.5 \\
PI     & 97  & 70  & 166.06  & 48.0 & 56.0 \\
HIPC   & 100 & 100 & 52.65   & 40.0 & 51.5 \\
DMCHBA & 100 & 100 & 4.88    & 22.5 & 70.0 \\
DGA & 100 & 100 & 1345.60 & 21.0 & 50.0 \\
\bottomrule
\end{tabular}

\vspace{4pt}
\parbox{\linewidth}{\footnotesize
\textit{Note:} Viable, stable, and $\Cterm$ summarize the primary pre-allocation comparison. MinMax and MinSum are route-length medians, in steps, over the 68 scenarios stable for all five methods and use the corresponding route objective.
}
\end{table}

Under ideal delivery, the primary all-attempt median $\Cterm$ values were 4.88~ms for DMCHBA, 52.65~ms for HIPC, 86.62~ms for ACBBA, 166.06~ms for PI, and 1.346~s for DGA. DMCHBA therefore approached DGA's MinMax route quality with more than two orders of magnitude lower full-protocol computation, while HIPC combined the second-lowest $\Cterm$ with closely competitive MinSum route quality.

\begin{figure}[!t]
\centering
\includegraphics[width=\columnwidth]{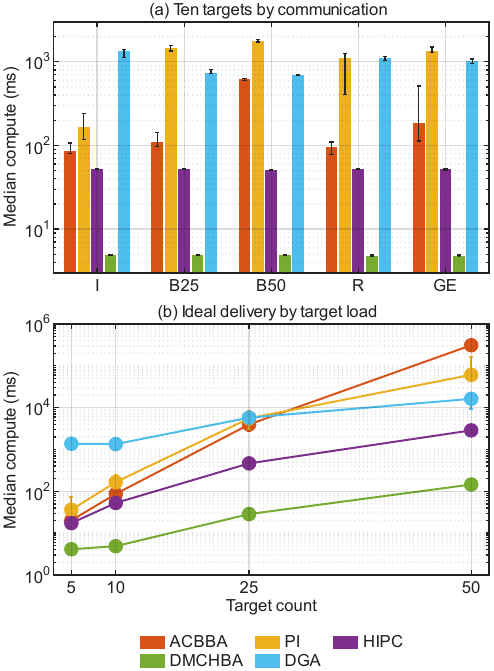}
\caption{Median full-protocol computation $\Cterm$. Panel (a) compares ten-target communication treatments, and panel (b) shows ideal-condition scaling with target count. Error bars are 95\% bootstrap confidence intervals. The vertical axis is logarithmic.}
\label{fig:prealloc_cterm}
\end{figure}

Across communication treatments, Fig.~\ref{fig:prealloc_cterm}(a) shows how the ten-target full-protocol computation budgets changed. DMCHBA remained near 4.8--4.9~ms across all five treatments, and HIPC remained near 51--53~ms. Under Bernoulli $p_{\mathrm{drop}}=0.50$, median $\Cterm$ increased to 618.84~ms for ACBBA and 1.758~s for PI, alongside stable counts of only 9 and 3. DGA's median fell to 698.31~ms, but only five trials became viable and four became stable. The lower terminal value reflected failed allocation availability rather than faster successful allocation.

For the primary comparison, the paired degraded-versus-ideal analysis found that ACBBA's $\Cterm$ increased significantly under Bernoulli $p_{\mathrm{drop}}=0.25$, Gilbert--Elliott loss, and Bernoulli $p_{\mathrm{drop}}=0.50$ (Holm-adjusted $p\leq.002$), but not under the Rayleigh treatment. PI showed a supported increase only under Bernoulli $p_{\mathrm{drop}}=0.50$ ($p=2.82\times10^{-5}$). DMCHBA showed no supported change. HIPC's supported changes were reductions smaller than 1.4~ms and did not alter its practical operating range. DGA decreased under every impaired treatment, but these reductions coincided with lower viable and stable rates.

\subsubsection{Effect of Target Load}

The broad route-quality ordering persisted as the static problem grew, with DGA and DMCHBA remaining closely matched on MinMax across all four target loads. For MinSum, DGA achieved the lowest median route cost at 5, 10, and 25 targets, while HIPC led at 50 targets. The load study therefore preserved the same leading methods while showing that their MinSum ordering can change as task count increases.

Reliability differences became more pronounced at higher loads. Under ideal delivery, HIPC, DMCHBA, and DGA remained viable and stable throughout the tested range, while ACBBA and PI increasingly separated viability from stability. Under Bernoulli $p_{\mathrm{drop}}=0.25$, HIPC and DMCHBA again remained viable and stable across all loads. DGA was reliable at the smaller problems but produced no viable allocation at 25 or 50 targets, while ACBBA and PI more often reached viable allocations without maintaining them through the stability criterion.

Full-protocol computation also diverged as target count increased, as shown in Fig.~\ref{fig:prealloc_cterm}(b). DMCHBA retained the lowest median $\Cterm$ at every load, with HIPC consistently second. ACBBA and PI grew much more steeply, while DGA remained substantially more expensive than DMCHBA despite preserving strong MinMax route quality. By 50 targets, median $\Cterm$ was 145.36~ms for DMCHBA and 2.863~s for HIPC, compared with 16.253~s for DGA, 60.794~s for PI, and 310.221~s for ACBBA. The load study therefore preserved the route-quality leaders while revealing increasingly large differences in allocation reliability and computational scaling.

\subsection{Execution-Integrated Computation}
\label{subsec:async_compute_results}

Because the static protocol and online execution accumulate work differently, the asynchronous timing study records every allocator invocation generated during complete ideal-communication CV missions. This produces a complementary measure of processor burden.

Cumulative allocator time depends on invocation frequency and the duration of each allocator invocation. At nominal ten-target load, CBAA achieved the lowest cumulative allocator computation at 13.60~ms per mission, with DMCHBA following at 25.15~ms and ACBBA a close third at 29.87~ms. DGA required 9.60~s, placing it as the most expensive by a large margin. To further decompose this metric, we also report allocator invocations. DMCHBA was called 66.61 times per mission, compared with 81.07 for CBAA, 108.20 for ACBBA, 116.36 for HIPC, 132.22 for PI, and 91.76 for DGA. DMCHBA therefore combined a moderate per-invocation cost with the lowest invocation frequency, while CBAA made more calls but required less time per invocation. PI combined the highest call count with a more expensive update, while DGA's evolutionary search made each triggered invocation substantially more expensive.

\begin{figure}[!t]
\centering
\includegraphics[width=\columnwidth]{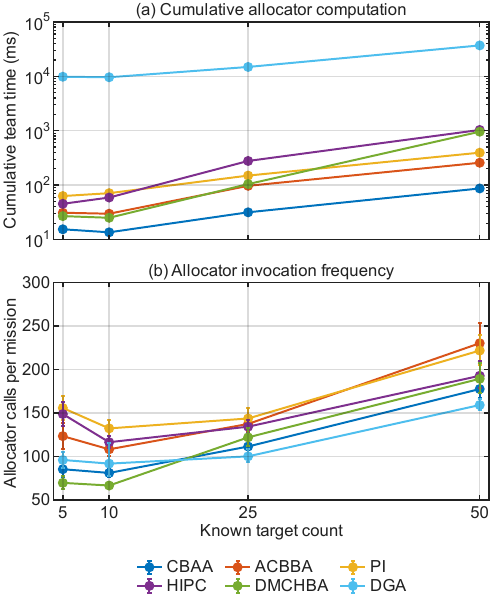}
\caption{Execution-integrated allocator computation under ideal communication using the primary MinMax settings. Panel (a) shows cumulative team allocator time and panel (b) allocator calls per mission across target loads. Error bars are 95\% paired-scenario bootstrap confidence intervals.}
\label{fig:async}
\end{figure}

As target load increased, algorithm ordering shifted. At 50 targets, CBAA still achieved the lowest cumulative team computation at 86.97~ms, while ACBBA and PI moved ahead of DMCHBA at 258.08~ms and 395.14~ms, respectively. The remaining times were 959.64~ms for DMCHBA, 1.03~s for HIPC, and 36.92~s for DGA. The five multi-task methods therefore had an asynchronous execution order of ACBBA, PI, DMCHBA, HIPC, and DGA, whereas their static $\Cterm$ order was DMCHBA, HIPC, DGA, PI, and ACBBA, showing a large shift in performance.

For the MinSum objective, HIPC, PI, and DGA required greater cumulative allocation time across the tested target loads. HIPC required roughly 1.8--3.1 times its MinMax allocation time, PI required roughly 1.9--2.4 times as much, and DGA required roughly 2.1--5.7 times as much. Allocator-call counts also increased for all three methods, showing that lower aggregate travel can require substantially more online allocation work.

Rank agreement between static $\Cterm$ and asynchronous cumulative mission computation was $\rho=0.9$ at five and ten targets, $0.5$ at 25 targets, and $-0.6$ at 50 targets. At ten targets, only ACBBA and HIPC exchanged positions between the static and asynchronous rankings. At larger loads, additional rank reversals appeared as execution-triggered allocation accumulated, particularly for DMCHBA and HIPC. Full-protocol static computation therefore characterizes allocation deliberation cost, while execution-integrated computation characterizes the processor work induced by the mission architecture.

\subsection{Scale Sensitivity}

Across the 32 measured grid--density--communication conditions, the broad MinMax ranking remained stable. DMCHBA and DGA achieved scale-wide mean ranks of 1.72 and 1.77 and mean deviations from the condition-best $\Smax$ of only 1.34\% and 1.19\%, respectively. At least one of DGA or DMCHBA occupied a top-two position in every condition, and both occupied the top two in 20 of 32 conditions, reflecting their MinMax performance in the main study. CBAA and HIPC formed the middle group with mean ranks of 3.19 and 3.61, while ACBBA and PI averaged 4.91 and 5.81 and did not enter the top two. 

\begin{figure}[htbp]
\centering
\includegraphics[width=\columnwidth]{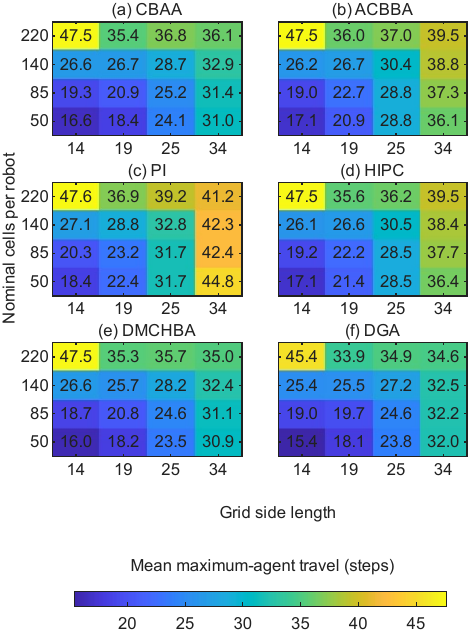}
\caption{Grid-size and robot-density sensitivity of mean realized MinMax travel $\Smax$ under ideal communication. Each cell contains ten targets, with $N_R=\lceil g^2/\delta\rceil$ robots. Annotations report means over retained complete six-algorithm blocks.}
\label{fig:grid_density_ideal}
\end{figure}

In the ideal-communication matrix of Fig.~\ref{fig:grid_density_ideal}, workspace size continued to affect the realized makespan proxy even when robot count increased with grid area. At 50 nominal cells per robot, the pooled mean $\Smax$ increased from 17.44 steps on the $14\times14$ grid with four robots to 35.36 steps on the $34\times34$ grid with 24 robots. At the sparsest robot setting, the $19\times19$ through $34\times34$ cases formed a broad 37--39-step saturation region. The one-robot $14\times14$ condition formed a separate limiting case in which communication and team deconfliction were irrelevant.

Within the fixed-ten-target design, increasing robot density produced diminishing MinMax reductions. The final team-size increment reduced pooled $\Smax$ by 13.8\% on the $14\times14$ grid but by only 1.5\% on the $34\times34$ grid. The curves flattened near four robots for the smallest grid, approximately five to eight robots for the intermediate grids, and approximately nine to fourteen robots for the largest grid. The Bernoulli $p_{\mathrm{drop}}=0.25$ conditions generally increased travel but preserved DGA and DMCHBA as the leading MinMax methods, supporting the same ranking conclusion and broad geometric pattern observed under ideal delivery.

A supporting MinSum analysis used 1,589 common complete six-algorithm blocks. DGA and HIPC remained the leading pair, with scale-wide mean total travel of 71.00 and 75.72 steps, respectively. DGA led 18 of 32 condition means and occupied the top two in all 32, while HIPC led 14 and occupied the top two in 30. The grid-density study therefore preserved the two core MinSum leaders, although their ordering shifted across the broader scale matrix.

\subsection{Implications for Allocator Selection}

HIPC and DGA form the leading MinSum operating region. HIPC retained the lowest impaired core mean at 66.95 steps, narrowly ahead of DGA at 67.22, while DGA led static MinSum route quality at three of four target loads and the grid-density aggregate. HIPC remained viable and stable throughout the pre-allocation study and retained substantially lower computational demand than DGA. Both methods achieved low aggregate travel with comparatively concentrated workload distributions, illustrating the tradeoff between MinSum performance, workload balance, reliability, and computation.

DGA and DMCHBA form the leading MinMax operating region. DGA achieved the lowest impaired mean at 24.49 steps and led 18 of the 24 impaired conditions, while DMCHBA followed closely at 24.78 steps and combined near-leading route quality with the lowest publication intensity and substantially lower computation. HIPC showed the strongest relative robustness under MinMax degradation, but DGA and DMCHBA retained better absolute MinMax performance through most of the sweep. This distinction separates degradation sensitivity from absolute route quality and makes DMCHBA particularly attractive when communication demand and computation matter alongside MinMax performance. Because computation was neither capped nor equalized across methods, the measured burden is best interpreted as part of each allocator’s observed route-quality tradeoff rather than as a shared experimental budget.

CBAA provides the lowest execution-integrated computation, but it has the highest publication intensity and remains outside the leading route-quality groups. ACBBA generally occupies an intermediate online position, while its static stability and full-protocol computation deteriorate with load and loss. PI is competitive for MinSum under ideal communication, but its third-place ideal result does not persist over the impaired aggregate, where it also has the highest duplicate-work mean. The preferred allocator therefore depends on route objective, communication environment, reliability requirements, workload-distribution tolerance, and the available static and online computation budgets.

\subsection{Relationship to Predecessor Benchmarks}

Nayak et al.\ found that ACBBA generally provided the strongest MinMax among their five methods, while PI or HIPC became more attractive when message demand received greater weight~\cite{2020nayak}. The larger paired suite used here shifts the leading MinMax group to DGA and DMCHBA. ACBBA occupied an intermediate position, while DMCHBA provided the strongest tradeoff between message demand and MinMax performance. This difference reflects the expanded algorithm set and execution framework rather than a direct contradiction of the earlier comparison. The present results also identify HIPC and DGA as the leading MinSum pair, an objective outside the primary outcome set of that benchmark.

Cao et al.\ identified DGA as the strongest method for conflict-free MinMax route quality and found that CBBA and HIPC often reached allocations with lower communication and computational cost~\cite{2025cao}. The present study reproduces DGA's MinMax strength, shows that DMCHBA approaches it with substantially lower full-protocol computation, and identifies HIPC and DGA as the leading MinSum methods across the present campaigns. It further distinguishes viable from stable allocation and shows that static computation ordering does not transfer unchanged to asynchronous execution.

\subsection{Limitations and Future Work}

Communication demand is measured through publication events containing standardized information density rather than serialized payload size, and measured computation is not inserted into simulated robot timing. These choices preserve a common communication abstraction across heterogeneous protocols and isolate allocator computation from the shared execution model, allowing differences to be attributed more directly to the allocation methods themselves. Additionally, the pre-allocation route costs report achieved solution quality rather than optimality gaps, while $\Cterm$ records full-protocol computation across both stable and round-capped attempts. Retaining capped attempts is important because failure to reach a stable allocation is itself part of the computational burden observed under degraded communication, but this metric combines computational efficiency with protocol outcome and duration. Future work will extend the benchmark through payload-level bandwidth measurements, compute-aware simulation, centralized MinMax and MinSum references, heterogeneous and obstacle-rich environments, dynamic task arrivals, and physical multi-robot validation.

\section{Conclusion}

This benchmark shows that decentralized MRTA selection under degraded communication depends jointly on route objective, communication demand, allocation reliability, and computation. DGA achieved the strongest average MinMax performance across the impaired asynchronous conditions. DMCHBA remained only 0.29 steps behind while producing the lowest publication intensity and requiring far less computation than DGA. HIPC and DGA formed the leading MinSum pair, with impaired means of 66.95 and 67.22 steps, respectively, and both occupied the top two in every impaired condition. Their MinSum performance was accompanied by greater workload concentration and online allocation cost. The grid-density study preserved DGA and DMCHBA as the leading MinMax pair, while DGA and HIPC remained the leading MinSum methods.

Static pre-allocation separated allocation availability, stability, route quality, and full-protocol computation. HIPC and DMCHBA remained viable and stable in every tested condition. DMCHBA had the lowest $\Cterm$ at every target load, while HIPC retained the second-lowest. DGA and DMCHBA produced the strongest complete MinMax routes, while DGA produced the lowest MinSum routes at 5, 10, and 25 targets and HIPC led at 50. During asynchronous execution, CBAA accumulated the least allocator computation. DMCHBA led the multi-task methods at smaller loads, while ACBBA led them at the two larger loads. The growing disagreement between static and execution-integrated rankings shows that isolated allocation runtime does not fully predict operational burden.

DGA is favored when route performance is prioritized and substantial computation is available. DMCHBA provides the strongest balance among MinMax performance, publication demand, and computation. HIPC provides closely competitive MinSum performance with stronger static allocation reliability and substantially lower computational demand than DGA, at the cost of concentrated workload distribution. PI remains competitive for MinSum under ideal communication, but that advantage does not persist through the full degradation suite. CBAA provides the lowest measured execution-integrated computation when processor work dominates communication and route-quality tradeoffs. The paired benchmark and common execution framework provide a basis for extending these choices to dynamic task arrivals, payload-aware communication, compute-induced control latency, heterogeneous teams, and physical multi-robot systems.

\balance
\bibliographystyle{IEEEtran}
\bibliography{references}

\end{document}